\documentclass[11pt]{article}
\usepackage[final]{acl}

\usepackage{times}
\usepackage{latexsym}

\usepackage[T1]{fontenc}

\usepackage[utf8]{inputenc}

\usepackage{microtype}

\usepackage{inconsolata}

\usepackage{graphicx}
\usepackage{booktabs}
\usepackage{array} 
\usepackage{multirow}
\usepackage{subcaption}
\usepackage{arydshln}
\usepackage{makecell}
\usepackage{amsmath}
\usepackage{tcolorbox}
\usepackage[dvipsnames]{xcolor}
\usepackage{tabularx}
\usepackage{amssymb}

\usepackage{csquotes}

\definecolor{todo}{rgb}{1,0,0}

\definecolor{ik}{rgb}{0,0,1}

\definecolor{care}{rgb}{0.961, 0.212, 0.259}

\usepackage{enumitem}

\usepackage{xcolor}
\usepackage{pifont}

\newcommand{\cmark}{\textcolor{green!65!black}{\ding{51}}}
\newcommand{\xmark}{\textcolor{red!65!black}{\ding{55}}}
\title{ReGround: Grounding Reviewer Comments in Multimodal Evidence}

\author{
Serwar Basch$^{\textbf{\texttt{1}}}$, Lizhen Qu$^{\textbf{\texttt{2}}}$, Iryna Gurevych$^{\textbf{\texttt{1}}}$
\\
        \textsuperscript{\textbf{\texttt{1}}}Ubiquitous Knowledge Processing Lab (UKP Lab), \\ Department of Computer Science 
and Hessian Center for AI (hessian.AI), TU Darmstadt \\
        \textsuperscript{\textbf{\texttt{2}}}Department of Data Science \& AI, Monash University, Australia \\
\href{https://www.ukp.tu-darmstadt.de}{www.ukp.tu-darmstadt.de}
}

\begin{document}
\maketitle

\begin{abstract}
Reviewer comments naturally relate to specific parts of the reviewed paper, yet grounding these comments to the underlying evidence is difficult due to long multimodal documents. Existing benchmarks do not capture this setting and largely focus on explicit, information-seeking queries. We introduce ReGround, a large-scale dataset for \emph{reviewer comment grounding} that links 10,267 reviewer comments to 16,274 evidence in the \emph{original anonymous submissions} of 3,656 papers. We build on a simple observation: author rebuttals often include explicit references to content of the submission used to address reviewer comments, providing a high-precision annotation source. We cast grounding as a retrieval task and evaluate a wide range of retrieval methods. Results show that retrieval over the entire paper content performs poorly, evidence-type inference is a major bottleneck, and multimodal evidence provides complementary signals that text alone misses. Our dataset exposes grounding reviewer comments as a difficult and practically important problem for scientific document understanding.\footnote{Our code and data are publicly available at \href{https://github.com/UKPLab/emnlp2026-reground}{https://github.com/UKPLab/emnlp2026-reground}}

\end{abstract}

\begin{figure}[t]
  \centering
  \includegraphics[width=\columnwidth]{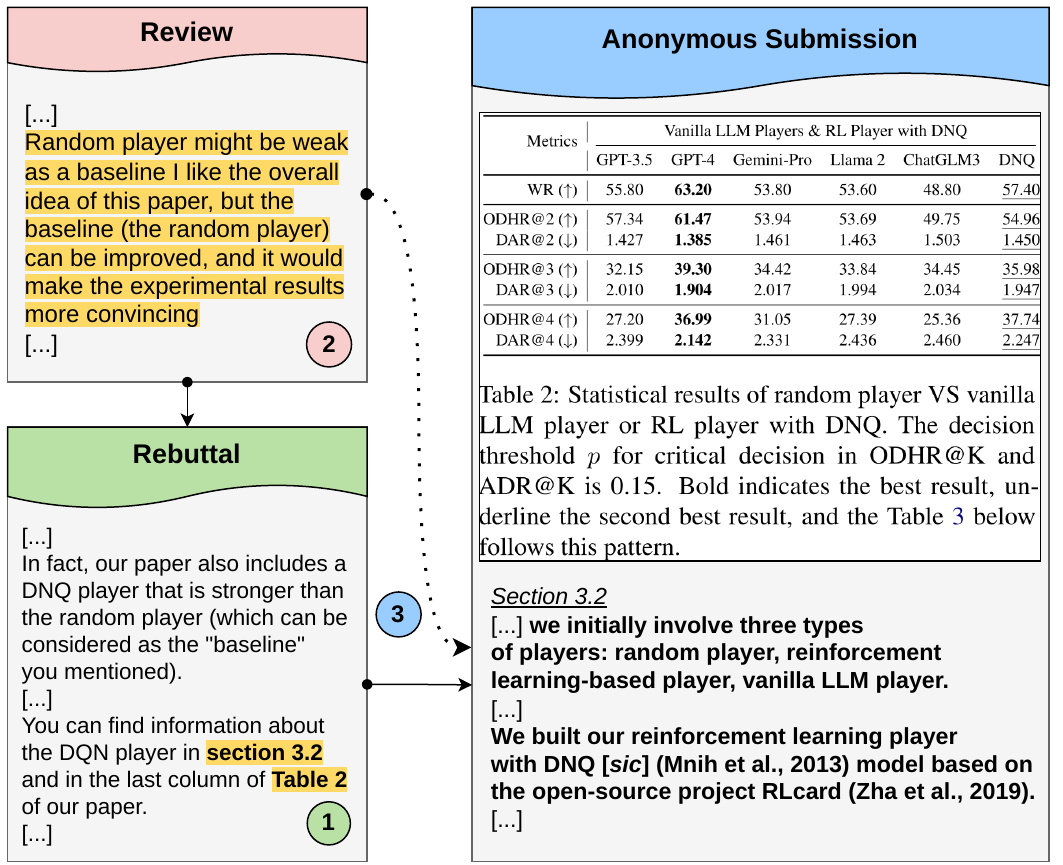}
  \caption{\textbf{Overview of dataset construction pipeline and a representative example}. We use explicit paper references in author rebuttals to derive high-precision links between reviewer comments and content in the original anonymous submission. (1) Detect rebuttal sentences that reference specific paper content. (2) Align each reference-bearing rebuttal sentence to the reviewer span it addresses. (3) Resolve the references in the submitted PDF, and extract the corresponding content, yielding a reviewer comment--paper content pair.}
  \label{fig:pipeline}
\end{figure}

\begin{table*}[!t]
\centering
\small
\setlength{\tabcolsep}{5pt}
\begin{tabular}{l c c c l l}
\toprule
\textbf{Dataset} & \textbf{Annotation Method} & \textbf{Original Sub.} & \textbf{Multimodal} & \textbf{Evidence Types} & \textbf{Size} \\
\midrule
% \textsc{NLPeer} & explicit-ref & \cmark & \xmark & \xmark & \xmark & paragraphs & 5672 papers \\
\textsc{ARIES} & manual & \xmark & \xmark & text spans & 42 papers \\
\textsc{PeerQA} & manual & \xmark & \xmark & sentences/paragraphs & 208 papers \\
\textsc{QASPER\footnotemark} & manual & \xmark & \cmark & paragraphs & 1585 papers \\
\textsc{F1000RD} & manual & \cmark & \xmark & lines/paragraphs/sections & 172 papers \\
\textsc{CLAIMCHECK} & manual & \cmark & \xmark & text spans & 41 papers \\
\midrule
\textbf{Ours} & \textbf{automatic} & \textbf{\cmark} & \textbf{\cmark} & \textbf{lines/sections/figures/tables} & \textbf{3656 papers} \\
\bottomrule
\end{tabular}
\footnotesize
\caption{Comparison of datasets that link queries and peer-review comments to full manuscript content.}
\label{tab:dataset_comparison}
\end{table*}

\section{Introduction}
\label{sec:introduction}
Peer review produces a large volume of expert comments about scientific papers, and the load is growing fast enough that top AI venues have begun deploying reviewer assistance systems into their official workflows~\citep{thakkar-etal-2025-review-feedback}. A core challenge for any such system is grounding a reviewer comment in the paper content it depends on: a methodological detail, a result table, a section-level claim, or a visual pattern in a figure. Beyond peer review, this instantiates a broader scientific-document problem: grounding natural, underspecified queries in long multimodal documents. The challenge is twofold: evidence is heterogeneous in modality, appearing as text, tables, or figures, and in granularity, from a single line to a full section.

%Peer review produces a large volume of expert comments about scientific papers. Many of these comments implicitly depend on specific paper content: a methodological detail, a result table, a section-level claim, or a visual pattern in a figure. A system that surfaces such evidence for a given comment could support more evidence-grounded reviewing, while also instantiating a broader scientific-document problem: grounding natural, underspecified queries in multimodal paper evidence. This problem is challenging because papers are long, heterogeneous, and multimodal, and because relevant evidence may appear as text, tables, figures, or multiple units at different granularities.

Existing resources do not directly support this setting (Table~\ref{tab:dataset_comparison}). Scientific QA datasets such as \textsc{QASPER} \citep{dasigi-etal-2021-dataset} and \textsc{PeerQA} \cite{baumgartner-etal-2025-peerqa} focus on information-seeking questions with relatively well-defined answer spans. Reviewer comments, by contrast, are often evaluative, underspecified, and implicitly connected to the paper content, as illustrated in Figure~\ref{fig:pipeline}. Revision-based datasets such as \textsc{ARIES} \cite{darcy-etal-2024-aries} link comments to edits in later versions of a paper rather than to the evidence available in the original submission. Finally, resources that target fine-grained links to the original paper, such as \textsc{F1000RD} \cite{kuznetsov-etal-2022-revise} and \textsc{CLAIMCHECK} \cite{ou-etal-2025-claimcheck}, are manually annotated, small, and text-only.

We identify author rebuttals as a high-precision source of grounding annotations. When responding to reviews, authors often explicitly point to specific parts of the submitted paper as evidence for addressing a concern, clarifying a point, or highlighting content that may have been overlooked. We refer to these mentions as \textit{paper references}, and the referenced content as \textit{evidence}. Because rebuttals discuss the paper as it was reviewed, these references expose links between reviewer comments and the original anonymous submission, rather than links to a revised or camera-ready version.\footnotetext{QASPER includes tables and figures (\textasciitilde11\%), but the analysis and experiments are limited to text only.}

On this basis, we introduce \texttt{\textbf{ReGround}}, a \textit{large-scale} dataset for grounding reviewer comments in multimodal evidence from scientific papers. \texttt{ReGround} contains 3,656 papers and 16,274 reviewer comment--evidence pairs covering paragraphs, sections, tables, and figures. \textbf{We formalize grounding as a retrieval task}: given a reviewer comment, the goal is to retrieve the evidence authors used to address it. Beyond peer review, the dataset targets a capability shared with scientific QA, fact-checking, and document-grounded assistance: retrieving multimodal evidence for natural-language queries over long papers. We benchmark sparse, dense, cross-encoder, LLM-based, and multimodal retrievers across text and visual modalities.

Our findings expose grounding as a problem with several distinct difficulties: (1) retrieval over a heterogeneous evidence pool is hard across all model families, with even the best LLM-based ranker reaching only 21\% Recall@10; (2) inferring the type of evidence (paragraph, section, figure, table) is a major bottleneck; (3) models also struggle with comments grounded to multiple evidence; (4) for figures and tables, visual and textual signals are complementary, with substantial modality-specific failure modes that motivate multimodal fusion rather than replacement.

%Our results show that (1) large language models (LLMs) outperform similarity-based approaches, suggesting that grounding often depends on more than surface-level similarity. (2) However, even strong LLMs fail to retrieve all relevant evidence, particularly in cases of multi-evidence.% when evidence consists of multiple types (e.g. a paragraph and a table).

%(3) We further find that performance improves substantially when the target evidence type is known in advance, highlighting the difficulty of selecting evidence from a heterogeneous pool. (4) Finally, multimodal representations yield consistent gains, indicating that different modalities provide complementary signals.

\section{Related Work}
\label{sec:related_work}

A growing line of work has introduced \textbf{peer-review corpora} to study the reviewing process and reviewer–author interactions \cite{kang-etal-2018-dataset, dycke-etal-2023-nlpeer, re2}. Several datasets further annotate review–rebuttal exchanges with discourse or argument structure, such as \textsc{APE} \cite{cheng-etal-2020-ape} and \textsc{DISAPERE} \cite{kennard-etal-2022-disapere}. Although valuable for modeling the review process, these corpora work on document level, while our work directly targets the fine-grained links between reviewer comments and paper content.

Other related datasets attempt to \textbf{link reviews with paper content}, but differ substantially from our setting. \textsc{ARIES} \cite{darcy-etal-2024-aries} and \textsc{Re3} \cite{ruan-etal-2024-re3} align reviewer comments with paper revisions, capturing how feedback manifests in revised papers rather than how comments relate to the original submission. \textsc{PeerQA} \cite{baumgartner-etal-2025-peerqa} reframes reviewer questions as document-level QA with manually annotated answer spans, but operates on camera-ready papers and is limited to text-only evidence.
More closely related, \citet{kuznetsov-etal-2022-revise} introduce the F1000RD dataset, and study implicit linking between review sentences and paper content, highlighting the difficulty of grounding underspecified review comments in long documents. However, their work is text-only and relies on manual annotation. Other efforts, such as \textsc{CLAIMCHECK} \cite{ou-etal-2025-claimcheck} and \textsc{ABCD-Link} \cite{basch2025}, focus on claim-centric or sentence-level linking and remain small in scale. Concurrent to our work, PRISMM-Bench \cite{selch2026prismmbench} compiles 384 reviewer-flagged multimodal inconsistencies from ICLR papers as a multiple-choice benchmark. In contrast, our dataset is derived automatically from author rebuttals, targets the original submission, scales to thousands of papers, and covers fine-grained multimodal evidence.

Our task is also related to \textbf{scientific evidence retrieval and document-level QA}, where systems must identify supporting evidence in long documents. Prior benchmarks include fact-checking datasets such as \textsc{SciFact} \cite{wadden-etal-2020-fact, wadden-etal-2022-scifact}, citation-span identification \cite{li-etal-2020-cist}, and QA over full papers such as \textsc{QASPER} \cite{dasigi-etal-2021-dataset}. Recent work has extended these settings to multimodal and PDF-based retrieval too \citep{dong-etal-2025-mmdocir}.
Our setting differs in several key respects: reviewer comments are naturally occurring not crowdsourced, are often evaluative rather than information-seeking; and figures and tables play a central role in addressing them. By grounding reviewer comments to fine-grained, multimodal evidence in original submissions, our dataset exposes a challenging and practically important retrieval problem that is not captured by existing benchmarks.

\section{Dataset Construction}
\label{sec:data_construction}

% We construct ReGround to support retrieval of multimodal evidence for a given reviewer comment. This requires preserving the original review context: the evidence must come from the submitted paper that authors and reviewers actually discuss. Our construction pipeline therefore combines review--rebuttal alignment, explicit reference detection, PDF-based content extraction, and filtering steps that remove cases where the reference does not point to evidence available in the original submission.

\subsection{Task Definition}
\label{subsec:task_definition}
We cast \emph{reviewer comment grounding} as a retrieval task. Given a reviewer comment $c$, the goal is to retrieve \emph{author-cited evidence}: paper content that addresses, clarifies, refutes, or contextualizes $c$. We build a candidate pool $D = \{d_1, \dots, d_n\}$ of evidence units extracted from the same original submission using a structured document representation (\S\ref{subsec:preprocessing}). A reviewer comment may be grounded in one or more units, so we evaluate both single- and multi-evidence retrieval (\S\ref{sec:experiments}).

Retrieval is a natural framing because it is the shared prerequisite for multiple downstream applications. We further target the \emph{implicit} setting, in which the comment does not itself contain an explicit reference (e.g., \enquote{Figure 3}, \enquote{Section 4.2}, \enquote{line 120}) to the gold evidence, since such references are trivially resolved by string matching.

\paragraph{Evidence units.}
The retrieval pool contains textual evidence units (paragraphs, sections, and figure/table captions) and visual evidence units (figure and table images). Line references are mapped to their containing paragraphs, because ACL-style line numbers are formatting artifacts rather than semantic units.
%We cast \emph{reviewer comment grounding} as a retrieval task. Given a reviewer comment \(c\), we retrieve paper content that grounds the comment, i.e., addresses, clarifies, or contextualizes the reviewer's comment. We refer to such content as \emph{evidence}. 

%We focus on retrieval because it is the shared prerequisite for downstream scientific-document applications. Furthermore, we target the implicit setting, where the reviewer comment itself does not already identify the evidence, because those cases can be trivially resolved with string matching.

%For a reviewer comment \(c\), we build a candidate pool \(D = \{d_1, \dots, d_n\}\) of evidence units extracted from the same original submission using a \textit{structured document representation} described below. %A reviewer comment may be grounded in one or more evidence units; evaluation may therefore consider both single-evidence and multi-evidence retrieval

%\paragraph{Evidence units.}
%The retrieval pool contains textual evidence units (paragraphs, sections, and figure/table captions), and visual evidence units (the extracted figure and table images). We map line references to their containing paragraphs, because ACL-style line numbers are formatting artifacts, not semantically meaningful units.%\footnote{This can be readily verified by inspecting individual lines in the present paper.}

\subsection{Data Source}
\label{subsec:data_source}

Our task requires the original submission to preserve the evidence that authors and reviewers actually discuss, in contrast to revised or camera-ready versions that may have been edited based on the reviews. To that end, we use version 2 of NLPeer \cite{dycke-etal-2023-nlpeer} with original submissions, reviews, and rebuttals from EMNLP~24/25, COLING~25, NAACL~25 and ACL~25.% We restrict our dataset to papers with at least one review–rebuttal pair.

\subsection{Preprocessing}
\label{subsec:preprocessing}

\paragraph{Review and rebuttal segmentation.}
We segment reviews and rebuttals into sentences using a custom rule-based splitter tailored to scientific writing. The splitter handles common reference patterns (e.g., ``Fig.~4'', ``Sec.~6.6'') and list-style formatting that breaks off-the-shelf tools.\footnote{See \S\ref{app:sentence_splitter} for implementation details.}

\paragraph{Paper content extraction.}
We parse the \emph{original submission} PDFs to create a structured document representation for each paper. Textual content is extracted at multiple levels: (i) individual numbered lines; (ii) section-level spans, identified via section headers and numbering; and (iii) figure and table captions. Since ACL-style papers follow standardized templates, we use PDF parsing via pypdf\footnote{\href{https://pypdf.readthedocs.io/en/stable/}{https://pypdf.readthedocs.io/en/stable/}} rather than OCR, because it ensures deterministic and consistent results. However, because paragraph boundaries cannot be reliably reconstructed from parsing PDFs, we use GROBID \citep{grobid} to parse them. Finally, we extract figures and tables as images along with their captions using PDFFigures~2.0 \citep{pdffigures}.

\subsection{Constructing Review-Evidence Links}
\label{subsec:constructing_gold_labels}

\paragraph{Detecting paper references.}
We identify rebuttal sentences that explicitly reference paper content using regular expressions developed iteratively over a random sampling of 250 rebuttals, and refined after a final manual inspection of the resulting dataset. The regex patterns cover references to lines, sections, tables, figures, appendices, equations, pages and footnotes.\footnote{See \S\ref{app:rebuttal_references} for details on the list of regular expressions.}

\paragraph{Aligning rebuttals to review comments.}
Authors often quote or index the reviewer comment they are addressing (e.g., ``W1'', ``Q5''). Using these cues, we employ an LLM (\texttt{gpt-oss-120b}\footnote{Chosen as a strong open-source model.}) to align each reference-bearing rebuttal sentence to its corresponding reviewer comment.\footnote{See \S\ref{app:llm_alignment} for prompt and setup details.} When a rebuttal addresses multiple reviewer segments, we concatenate them. Manual evaluation of alignment quality is detailed in \S\ref{subsec:qc}.

\paragraph{Resolving paper references.}
We resolve each detected paper reference against the structured document representation. Line references are mapped to the paragraphs that contain them, section references are matched by number and title, and figure/table references are matched by index and linked to both their captions and extracted images.

\subsection{Filtering}
\label{subsec:filtering}

The raw links extracted from rebuttals contain three sources of noise, which we remove with successive filters (see \S\ref{app:implicit_filtering} for details). First, authors frequently quote reviewer text verbatim, which can trigger false-positive reference detection; we discard rebuttal sentences that overlap with the review using fuzzy string matching (\textasciitilde20\% of sentences). Second, to focus on the implicit setting, if a reviewer comment already references the same paper content referenced in the rebuttal, grounding becomes trivially solvable with string matching, so we drop such pairs (\textasciitilde18\% of pairs). Third, we prompt \texttt{gpt-oss-120b} to remove pairs where the author-cited evidence is not actually present in the original submission\footnote{See \S\ref{app:implicit_filtering} for implementation details.}, namely (i) promised camera-ready changes, (ii) new experiments introduced in the rebuttal, or (iii) content from external papers (\textasciitilde23\% of pairs); we validate this step manually in \S\ref{subsec:qc}.

\begin{table}[!t]
\centering
\begin{tabular}{l r}
\toprule
\textbf{Statistic} & \textbf{Value} \\
\midrule
Papers & 3656 \\
Reviewer comments (queries) & 10267 \\
Paper references (evidence) & 16274 \\
Unique Paper references & 13391 \\
\midrule
Avg. queries per paper & 2.81 \\
Avg. evidence per query & 1.58 \\
\% Queries with $\geq$ 2 evidence  & 25.77 \\
\bottomrule
\end{tabular}
\caption{Summary statistics of the final dataset.}
\label{tab:dataset_stats}
\end{table}

\begin{figure*}[!t]
    \centering
    \includegraphics[width=\linewidth]{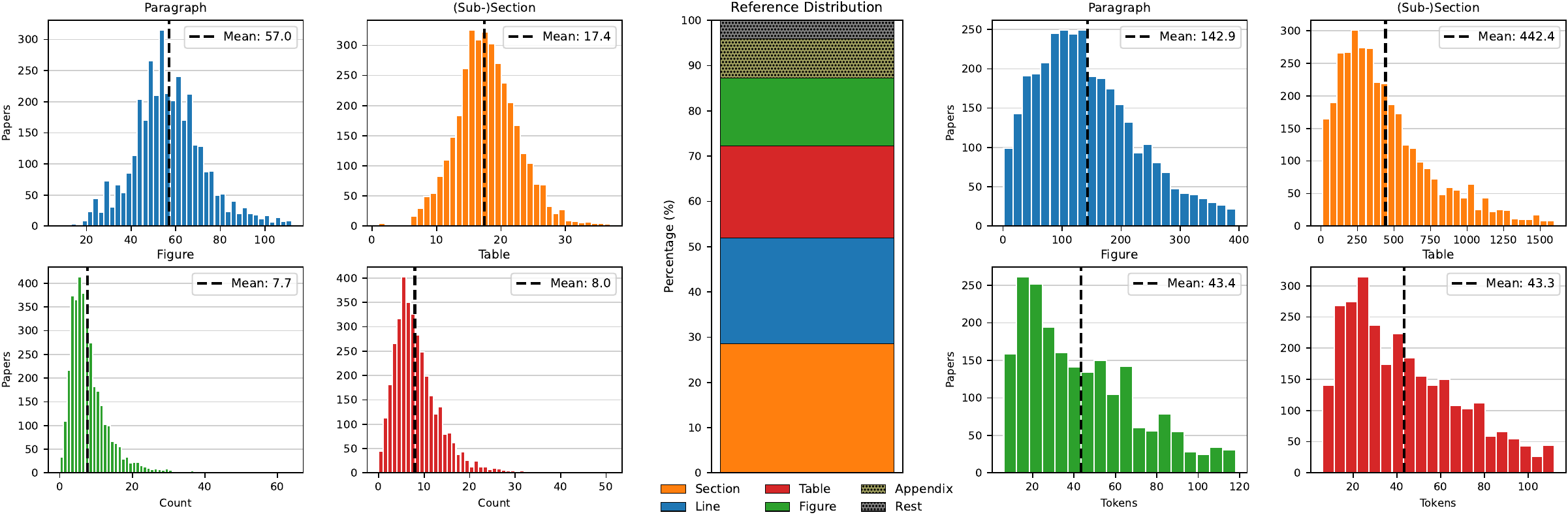}
    \caption{Dataset composition and evidence types. \textit{Left:} Distribution of number of candidate units of each type (paragraphs, sections, figures, tables) per paper, showing substantial variation in retrieval pool size across types. \textit{Center:} Distribution of reference types in the dataset. Rest includes references to pages, footnotes, and equations. \textit{Right:} Token-length distributions of textual evidence units by type, illustrating large disparities in semantic scope.}

    \label{fig:dataset_stats}
\end{figure*}

\subsection{Quality Assurance}
\label{subsec:qc}
%We conduct manual evaluations to assess both the \emph{review--rebuttal alignment} and the \emph{LLM-filtering} used in dataset construction. We also do a manual inspection of the final dataset.

We manually evaluate the two LLM-based stages of the pipeline and inspect the final dataset.

\paragraph{Review--rebuttal alignment.}
Two annotators\footnote{A graduate CS student and a co-author of this paper.} verify alignments for 100 randomly sampled papers, achieving 96\% alignment acceptance (with Cohen's $\kappa$=0.83). Most prevelant errors (13\%) are under-selection of the reviewer comment span rather than incorrect alignment.

\paragraph{Filtering validation.}
%We evaluate the LLM-filtering by annotating 300 reviewer comment--rebuttal sentence pairs using the same criteria given to the model (Cohen's $\kappa = 0.84$). Compared against this gold subset, the LLM-filtering achieves 97\% precision and 93\% recall, reflecting a small bias toward removing ambiguous cases. The high precision is expected, as the filtering criteria rely on simple surface-level cues, including references to future revisions (e.g., ``we will update...''), mentions of new experiments introduced in the rebuttal, and standard citation patterns for external content.

We annotate 300 reviewer comment–rebuttal sentence pairs using the same criteria (\S\ref{subsec:filtering}) given to the model ($\kappa = 0.84$). The LLM filter achieves 97\% precision and 93\% recall against this gold subset, with a small bias toward removing ambiguous cases. The high precision is expected since the filtering criteria rely on surface-level cues such as future-tense references ("we will update..."), mentions of new experiments, and standard citation patterns.\footnote{See \S\ref{app:manual_validation} for details on the annotation setup.}

\paragraph{Manual inspection.}
%Manual inspection of 200 reviewer comments shows that our pipeline is reliable: A small fraction (1.58\%) of evidence spans may be coarse-grained (i.e. the author refer to a whole section when a subsection would be more precise), though determining these boundaries is often subjective, so we decide to keep them in the dataset, because they represent realistic author-reviewer interactions. For a subet of 100 comments, we also check whether the referenced evidence is genuinely valid, and find out that all evidence is topically relevant. More importantly, in the majority of cases (84\%) the reviewers themselves acknowledged the rebuttal, so it is safe to assume that the provided evidence was accepted by them too.\footnote{See \S\ref{app:appendix_manual_validation} for details on the manual inspection.}

We inspect 200 reviewer comments and their 316 referenced evidence to assess reference resolution and evidence validity. We find that 1.58\% of evidence spans are coarse-grained (e.g., a whole section cited when a subsection would suffice); we retain these since they reflect realistic author behavior and the appropriate boundary is often subjective. On a 100-comment subset, we find that all referenced evidence is relevant to the reviewer comment, and in 84\% of cases the reviewer explicitly acknowledged the rebuttal, indirectly validating their relevancy. We note that this acknowledgment represents the reviewer being satisfied with the rebuttal, and not a signal that the review was refuted.\footnote{See \S\ref{app:appendix_manual_validation} for details on the manual inspection.}

\subsection{Dataset Statistics}
\label{subsec:stats}

\paragraph{Dataset scale.}
Our final dataset contains 3,656 papers, comprising 10,267 reviewer comments and 16,274 paper evidence (Table~\ref{tab:dataset_stats}). Notably, 25.77\% of comments have multiple evidence, underscoring the need for methods that can effectively handle multi-evidence scenarios.

\paragraph{Evidence types.}
Figure~\ref{fig:dataset_stats} shows that references to lines, sections, figures, and tables account for nearly 90\% of all evidence, motivating our focus on them, while appendix references comprise around 8\%. The distribution of evidence varies substantially by type, and notably, textual evidence varies widely in length, from short captions to long section-level spans, resulting in uneven retrieval pools across evidence types. This complicates retrieval, as models must retrieve from candidate units that differ in length and information density. 

\paragraph{Limits of semantic similarity.}
\label{para:semantic_gap}
To quantify the relation between reviewer comments and their gold evidence, we compute their cosine similarity and obtain an average of 0.377, indicating limited semantic overlap. For comparison, the similarity between the corresponding rebuttal sentence and gold evidence is 0.431. This is expected: reviewer comments are often evaluative and underspecified, and the referenced evidence frequently corresponds to content that reviewers may have missed or overlooked. This weak similarity suggests that retrieval based on semantic matching alone is insufficient, motivating the need for better methods.\footnote{See \S\ref{app:similarity} for details on the distribution of cosine similarity.}

\paragraph{Types of reviewer comments.}
Finally, we manually annotate 200 randomly sampled comments with a taxonomy of addressed paper issues. The most common category is Empirical Rigor (31\%), encompassing concerns about experimental design, evaluation, and analysis. Two other categories--Technical Soundness and Deployment \& Impact--each account for \textasciitilde16\%. This distribution confirms that the dataset captures a broad range of reviewer concerns grounded in diverse parts of the paper.\footnote{See \S\ref{app:taxonomy} for the taxonomy results and explanation.}

\subsection{Use Cases}
\label{sec:use-cases}

ReGround supports reviewer-facing systems that connect review comments to the specific parts of a paper they concern. First, in a human reviewer-assistance workflow, a system can retrieve relevant paper locations for a drafted comment. This helps reviewers verify a claim against the submission, identify potentially overlooked material, and attach precise citations to the review. The reviewer remains responsible for the final judgment, while retrieval reduces the effort required to check and ground comments.

Second, ReGround provides supervision and evaluation data for LLM-based reviewer systems. Such systems should not only generate criticisms or questions, but also link them to concrete passages, sections, figures, and tables in the paper. This grounding can make generated reviews more transparent, easier for humans to verify, and less likely to rely on unsupported claims.

\section{Experimental Setup}
\label{sec:experiments}

\texttt{\textbf{ReGround}} supports tasks that require grounding natural-language inputs in scientific paper evidence (review assistance, scientific QA, fact-checking), where retrieval of relevant evidence is the shared prerequisite. We therefore evaluate grounding as retrieval, and study research questions that isolate the different aspects making it hard:

\begin{itemize}[leftmargin=*,topsep=0pt,parsep=0pt,partopsep=0pt]
    \item \textbf{RQ1:} How well do models retrieve over a heterogeneous textual evidence pool, when the evidence type is unknown?%How do models perform when retrieving over a heterogeneous textual evidence pool without knowing the evidence type?
    \item \textbf{RQ2:} How well can models retrieve when the evidence type is known?
    \item \textbf{RQ3:} How well do models retrieve \emph{multiple} non-redundant evidence units?%How do models perform when retrieving \emph{all} relevant evidence units at once?
    \item \textbf{RQ4:} How much does visual information add to grounding figures and tables, beyond captions?
\end{itemize}

\subsection{Evaluation Settings}
\label{subsec:eval_settings}

\paragraph{Unified text retrieval (RQ1).}
We retrieve from a single heterogeneous \emph{textual} candidate pool containing paragraphs, sections, and figure/table captions. In this setting, models must both identify \emph{what} content is relevant and infer \emph{which evidence type} it appears in. Evaluation is performed for each referenced evidence individually then averaged over all comment-evidence pairs.

\paragraph{Type-aware (oracle) text retrieval (RQ2).}
To separate retrieval difficulty from type inference, we define an oracle setting restricted to the gold evidence type. For example, comments grounded to sections are evaluated against section candidates only, and comments grounded to tables against table captions. This setting quantifies the performance loss due to heterogeneous candidate pools.

\paragraph{Type-aware joint evidence retrieval (RQ3).}
To directly test multi-evidence grounding, we additionally evaluate a joint setting on comments with \(\geq 2\) grounded targets. Here, models are evaluated on their ability to retrieve the full set of relevant evidence. We assume oracle type information as above to focus on the multi-evidence challenge rather than type inference.

\paragraph{Visual evidence retrieval (RQ4).}
In this setup, we use figures and tables represented as \emph{images}. We evaluate two image representations: (i) the raw figure/table image alone, and (ii) the image augmented with its caption, where the caption text is concatenated visually to the image input.

\subsection{Baselines}
\label{subsec:models}
\paragraph{Text retrieval models.}
We use a range of retrieval models: Sparse baselines (\texttt{BM25} \cite{INR-019} and \texttt{SPLADEv3} \cite{lassance2024spladev3newbaselinessplade}). Dense encoders (\texttt{all-mpnet-base-v2}\footnote{\href{https://huggingface.co/sentence-transformers/all-mpnet-base-v2}{sentence-transformers/all-mpnet-base-v2}}\footnote{Abbreviated further as \texttt{\textbf{all-mpnet}} for brevity}, \texttt{BGE-M3} \cite{chen-etal-2024-m3}, \texttt{Qwen3-Embedding-4B} \cite{zhang2025qwen3embeddingadvancingtext}, and \texttt{EmbeddingGemma} \cite{vera2025embeddinggemmapowerfullightweighttext}) which independently encode comments and evidence and rank by cosine similarity. To assess joint query--document modeling, we also include cross-encoders: \texttt{ms-marco-MiniLM-L12-v2}\footnote{\href{https://huggingface.co/cross-encoder/ms-marco-MiniLM-L12-v2}{cross-encoder/ms-marco-MiniLM-L12-v2}}\footnote{Abbreviated further as \texttt{\textbf{MiniLM-L12}} for brevity} and \texttt{bge-reranker-v2-m3} \cite{chen-etal-2024-m3}.

\paragraph{LLM-based ranking.}
We evaluate LLMs as pointwise relevance scorers. Given a reviewer comment and a candidate evidence, the model outputs a binary relevance judgment (Yes/No). We convert the binary decision to a continuous relevance score using a two-class softmax over the decision tokens, $s = \exp(\ell_{\texttt{Yes}}) / (\exp(\ell_{\texttt{Yes}}) + \exp(\ell_{\texttt{No}}))$, where $\ell_{\texttt{Yes}}$ and $\ell_{\texttt{No}}$ are the model-assigned token log-probabilities. We evaluate non-reasoning LLMs: \texttt{Gemma~3 (12B, 27B)} \cite{gemmateam2025gemma3technicalreport}, \texttt{Qwen~3~ (4B, 30B Instruct)} \cite{yang2025qwen3technicalreport}, and reasoning LLMs: \texttt{Qwen~3~30B Thinking} \cite{yang2025qwen3technicalreport} and \texttt{gpt-oss~20b} \cite{openai2025gptoss120bgptoss20bmodel}, and \texttt{GPT-5.1}\footnote{gpt-5.1-2025-11-13} as a representative commercial model. Due to cost, \texttt{GPT-5.1} is evaluated on a random 50\% subset.\footnote{See \S\ref{app:llm_ranking} for implementation details.}
%All models use the same zero-shot prompt and scoring protocol

\paragraph{Visual retrieval models.}
For image-based retrieval, we evaluate vision--text encoders including \texttt{SigLIP2} \cite{siglip2}, \texttt{OpenCLIP} \cite{openclip}, and \texttt{Jina Embeddings v4} \cite{jina}, which rank candidates by cosine similarity. We also evaluate a late-interaction model, \texttt{ColQwen2.5}, and evaluate multimodal LLMs (\texttt{Qwen~3~VL 8B, 32B} \cite{yang2025qwen3technicalreport}) using the same scoring protocol.

\begin{table}[t]
\centering
\small
\renewcommand{\arraystretch}{1}
\resizebox{\linewidth}{!}{
\begin{tabular}{lcccc}
\toprule
\textbf{Model} & \textbf{MRR} & \textbf{R@1} & \textbf{R@2} & \textbf{R@10} \\
\midrule
\textit{\textbf{Sparse encoders}} & & & & \\
\texttt{BM25} & \texttt{7.62} & \texttt{3.57} & \texttt{5.65} & \texttt{14.90} \\
\texttt{SPLADEv3} & \texttt{7.39} & \texttt{3.04} & \texttt{5.38} & \texttt{15.17} \\
\midrule
\textit{\textbf{Dense encoders}} & & & & \\
\texttt{all-mpnet} & \texttt{9.08} & \texttt{4.08} & \texttt{6.78} & \texttt{19.12} \\
\texttt{BGE-M3} & \texttt{9.41} & \texttt{4.45} & \texttt{7.14} & \texttt{19.05} \\
\texttt{Qwen-3 Embedding 4B} & \texttt{6.92} & \texttt{3.26} & \texttt{5.03} & \texttt{14.59} \\
\texttt{EmbeddingGemma} & \texttt{8.78} & \texttt{4.17} & \texttt{6.64} & \texttt{17.68} \\
\midrule
\textit{\textbf{Cross encoders}} & & & & \\
\texttt{BGE-M3-Reranker} & \texttt{7.85} & \texttt{3.42} & \texttt{5.67} & \texttt{16.20} \\
\texttt{MiniLM-L12} & \texttt{7.92} & \texttt{3.62} & \texttt{5.82} & \texttt{15.81} \\
\midrule
\textit{\textbf{LLMs}} & & & & \\
\texttt{Gemma-3 27B} & \texttt{10.12} & \texttt{4.68} & \texttt{7.74} & \texttt{20.34} \\
\texttt{Qwen-3 30B Instruct} & \textbf{\texttt{10.87}} & \textbf{\texttt{5.12}} & \textbf{\texttt{8.29}} & \textbf{\texttt{21.15}} \\
\bottomrule
\end{tabular}}
\caption{Unified text retrieval performance. Retrieval is performed over a heterogeneous pool of textual evidence (paragraphs, sections, and captions) without evidence type hints.}
\label{tab:unified_text_retrieval}
\end{table}

\begin{table*}[!t]
\centering
\renewcommand{\arraystretch}{1}
\resizebox{\textwidth}{!}{
\begin{tabular}{lccccccc ccc ccc}
\toprule
 & \multicolumn{4}{c}{\textbf{Paragraph}} 
 & \multicolumn{3}{c}{\textbf{Section}} 
 & \multicolumn{3}{c}{\textbf{Table Caption}} 
 & \multicolumn{3}{c}{\textbf{Figure Caption}} \\
\cmidrule(lr){2-5} \cmidrule(lr){6-8} \cmidrule(lr){9-11} \cmidrule(lr){12-14}
\textbf{Model} 
& \textbf{MRR} & \textbf{R@1} & \textbf{R@2} & \textbf{R@10}
& \textbf{MRR} & \textbf{R@1} & \textbf{R@2}
& \textbf{MRR} & \textbf{R@1} & \textbf{R@2}
& \textbf{MRR} & \textbf{R@1} & \textbf{R@2} \\
\midrule
\textit{\textbf{Sparse encoders}} & & & & & & & & & & & & &\\
\texttt{BM25} 
& \texttt{18.65} & \texttt{8.11} & \texttt{13.37} & \texttt{35.58}
& \texttt{25.55} & \texttt{11.38} & \texttt{19.58}
& \texttt{47.28} & \texttt{24.38} & \texttt{41.10}
& \texttt{44.25} & \texttt{22.57} & \texttt{38.65} \\

\texttt{SPLADEv3} 
& \texttt{27.92} & \texttt{15.13} & \texttt{23.44} & \texttt{48.89}
& \texttt{29.81} & \texttt{15.24} & \texttt{24.54}
& \texttt{53.51} & \texttt{30.77} & \texttt{49.14}
& \texttt{48.59} & \texttt{26.51} & \texttt{44.69} \\

\midrule
\textit{\textbf{Dense encoders}} & & & & & & & & & & & & &\\
\texttt{all-mpnet} 
& \texttt{19.38} & \texttt{8.39} & \texttt{13.29} & \texttt{36.81}
& \texttt{28.98} & \texttt{14.02} & \texttt{23.70}
& \texttt{49.50} & \texttt{26.75} & \texttt{44.12}
& \texttt{48.17} & \texttt{27.07} & \texttt{43.26} \\

\texttt{BGE-M3}
& \texttt{24.38} & \texttt{12.63} & \texttt{18.91} & \texttt{42.59}
& \texttt{27.55} & \texttt{13.25} & \texttt{21.62}
& \texttt{51.61} & \texttt{28.99} & \texttt{45.80}
& \texttt{49.05} & \texttt{26.43} & \texttt{46.00} \\

\texttt{Qwen-3 Embedding 4B} 
& \texttt{14.54} & \texttt{7.13} & \texttt{11.02} & \texttt{25.93}
& \texttt{22.87} & \texttt{12.10} & \texttt{19.58}
& \texttt{39.62} & \texttt{22.78} & \texttt{36.22}
& \texttt{37.10} & \texttt{21.32} & \texttt{35.20} \\

\texttt{EmbeddingGemma}
& \texttt{27.11} & \texttt{14.08} & \texttt{22.35} & \texttt{47.55}
& \texttt{31.76} & \underline{\texttt{17.31}} & \texttt{26.69}
& \texttt{53.25} & \texttt{30.15} & \texttt{48.99}
& \texttt{50.91} & \texttt{28.49} & \texttt{47.98} \\

\midrule
\textit{\textbf{Cross encoders}} & & & & & & & & & & & & &\\
\texttt{BGE-M3-Reranker}
& \texttt{27.05} & \texttt{14.58} & \texttt{22.05} & \texttt{46.57}
& \texttt{29.32} & \texttt{14.69} & \texttt{24.02}
& \texttt{53.86} & \texttt{30.92} & \texttt{49.83}
& \texttt{50.65} & \texttt{28.19} & \texttt{48.26} \\

\texttt{MiniLM-L12}
& \texttt{24.38} & \texttt{13.25} & \texttt{19.85} & \texttt{41.94}
& \texttt{28.48} & \texttt{14.39} & \texttt{22.60}
& \texttt{48.51} & \texttt{26.26} & \texttt{42.67}
& \texttt{47.19} & \texttt{25.54} & \texttt{42.85} \\

\midrule
\textit{\textbf{LLMs}} & & & & & & & & & & & & &\\
\texttt{Gemma-3 12B} 
& \texttt{32.55} & \texttt{18.73} & \texttt{27.72} & \texttt{53.05}
& \texttt{30.76} & \texttt{14.87} & \texttt{26.15}
& \texttt{53.41} & \texttt{30.89} & \texttt{49.89}
& \underline{\texttt{53.20}} & \texttt{32.59} & \underline{\texttt{50.68}} \\

\texttt{Gemma-3 27B} 
& \underline{\texttt{34.87}} & \underline{\texttt{20.32}} & \underline{\texttt{30.01}} & \underline{\texttt{57.62}}
& \texttt{31.36} & \texttt{14.48} & \texttt{26.37}
& \texttt{53.72} & \texttt{31.65} & \texttt{49.87}
& \texttt{52.25} & \texttt{32.50} & \texttt{49.00} \\

\texttt{Qwen-3 4B}
& \texttt{31.08} & \texttt{18.02} & \texttt{27.17} & \texttt{54.16}
& \texttt{30.49} & \texttt{14.47} & \texttt{25.12} & \texttt{54.31}
& \texttt{31.59} & \texttt{49.95} & \texttt{52.01} & \texttt{31.30} & \texttt{48.80} \\

\texttt{Qwen-3 30B Instruct} 
& \texttt{32.97} & \texttt{19.19} & \texttt{27.69} & \texttt{53.64}
& \textbf{\texttt{33.88}} & \textbf{\texttt{18.16}} & \textbf{\texttt{30.46}}
& \textbf{\texttt{56.17}} & \textbf{\texttt{33.31}} & \textbf{\texttt{53.29}}
& \textbf{\texttt{55.40}} & \textbf{\texttt{34.37}} & \textbf{\texttt{55.16}} \\

\texttt{Qwen-3 30B Thinking} 
& \texttt{25.44} & \texttt{13.76} & \texttt{20.76} & \texttt{42.39}
& \texttt{30.26} & \texttt{15.20} & \texttt{25.05}
& \texttt{50.91} & \texttt{28.92} & \texttt{44.87}
& \texttt{46.46} & \texttt{25.11} & \texttt{42.31} \\

\texttt{gpt-oss-20b} 
& \texttt{28.36} & \texttt{15.71} & \texttt{24.44} & \texttt{44.14}
& \texttt{29.39} & \texttt{14.89} & \texttt{24.32}
& \texttt{47.30} & \texttt{24.36} & \texttt{40.83}
& \texttt{45.29} & \texttt{24.39} & \texttt{41.01} \\

\texttt{GPT-5.1} 
& \textbf{\texttt{36.26}} & \textbf{\texttt{21.62}} & \textbf{\texttt{31.34}} & \textbf{\texttt{59.07}}
& \underline{\texttt{32.98}} & \texttt{16.51} & \underline{\texttt{28.73}}
& \underline{\texttt{54.26}} & \underline{\texttt{31.69}} & \underline{\texttt{50.63}}
& \texttt{52.64} & \underline{\texttt{32.86}} & \texttt{49.86} \\

\bottomrule
\end{tabular}}
\caption{Type-aware (oracle) text retrieval performance. Retrieval is restricted to the gold evidence type, isolating retrieval difficulty from type inference. Note that \texttt{GPT-5.1} is evaluated on a 50\% subset due to cost. Cutoff for figures and tables is at 2 because of the smaller pool size, and for sections at 2 because higher cutoffs correspond to more than one page which is impractical.}
\label{tab:oracle_text_retrieval}
\end{table*}

\subsection{Evaluation Metrics}
\label{subsec:metrics}
We report \textbf{Recall@\(k\)} to measure the fraction of the relevant set in the top-\(k\) ranked candidates averaged across queries, and Mean Reciprocal Rank (\textbf{MRR}) to measure how highly the first relevant unit is ranked, emphasizing early precision.

We evaluate at the level of evidence units with exact unit-ID matching: a prediction is counted as correct only when it matches the gold unit exactly, and hierarchically related units (e.g., a section containing the gold paragraph) receive no partial credit. Each reference resolves to a single evidence unit to avoid inflating scores via ancestor units.

\section{Results}
\label{sec:results}
\subsection{Unified Text Retrieval (RQ1)}
Table~\ref{tab:unified_text_retrieval} shows that unified retrieval over a heterogeneous textual pool is difficult across all model families: MRR remains below 11 and R@10 peaks at 21.15. LLM-based rankers achieve the best results, but the gains over the dense models are modest (10.87 vs. 9.41 MRR), indicating substantial remaining headroom. More notably, cross-encoders fail to outperform bi-encoders despite their stronger joint query–document modeling, a counterintuitive result that motivates a closer look. 

We analyze the evidence length effect in detail in \S\ref{app:length_effect}, and find that cross-encoders (\texttt{BGE-M3-Reranker}) collapse on the longest section quantile, with MRR dropping from 44 at Q5 to near zero at Q10, while LLM-based rankers continue to improve up to Q8 and degrade only gracefully thereafter. Since the unified pool mixes short captions with long section-level spans (cf. Figure~\ref{fig:dataset_stats}), cross-encoders are penalized on these instances. These results establish the difficulty of unified text retrieval, and motivate model choices that are robust to long, heterogeneous candidates.

\subsection{Type-aware (oracle) Text Retrieval (RQ2)}
Table~\ref{tab:oracle_text_retrieval} shows that restricting retrieval to the gold evidence type substantially improves performance across all models and evidence types. These performance gains indicate that evidence-type uncertainty is a major source of error in the unified setting.

%Relative to unified retrieval, the large gains indicate that evidence-type uncertainty is a major source of error in the unified setting.

Nonetheless, performance still varies by type. Paragraph- and section-level retrieval remain challenging: even the strongest models achieve only around 31\% Recall@2, consistent with the observation of semantic gap between reviewer comments and gold evidence (\S\ref{para:semantic_gap}). At higher cutoffs, recall improves, but retrieving 10 paragraphs can correspond to more than a page of text in some papers, which reduces its practical usefulness.%\footnote{We set cutoff for figures and tables at 2 because of the smaller pool size, and for sections at 2 because higher cutoffs correspond to more than one page which is impractical.}

Caption retrieval achieves the highest scores overall. However, caption candidate pools are also smaller than paragraph/section pools, which likely contributes to the higher recall (cf. Figure~\ref{fig:dataset_stats}).

Across model families, larger LLMs perform best under oracle conditions, but the gap between the models narrows compared to unified retrieval. This pattern suggests that improving routing to the correct evidence type is an essential bottleneck.

We also test whether this oracle gain can be recovered by predicting the evidence type before retrieval. Concretely, hard filtering by predicted type drops \texttt{Qwen-3 30B Instruct's} R@10 from 21.15 to 6.11 (\S\ref{app:predicted_type_routing}), because errors on figure/table prediction remove the correct evidence from the pool. Type prediction is thus better viewed as a soft reranking signal than a hard filter, a finding that shapes how systems should integrate routing.

\subsection{Type-aware joint evidence retrieval (RQ3)}
%Table~\ref{tab:joint_gold_label_retrieval_1} shows the best model performances when retrieving the entire referenced units set at once. The performance remains limited at small cutoffs with models retrieving only a small fraction of gold evidence. This suggests that multi-evidence grounding introduces a distinct challenge beyond type inference, likely because relevant support is distributed across multiple non-redundant units.

\begin{table}[!t]
\centering
\small
\renewcommand{\arraystretch}{1}
\resizebox{\linewidth}{!}{
\begin{tabular}{lcccc}
\toprule
\textbf{Model} & \textbf{MRR} & \textbf{R@1} & \textbf{R@2} & \textbf{R@10} \\
\midrule
\texttt{EmbeddingGemma} & \texttt{37.63} & \texttt{10.25} & \texttt{18.79} & \texttt{55.94} \\
\midrule
\texttt{Qwen-3 30B Instruct} & \texttt{\textbf{43.29}} & \texttt{\textbf{12.69}} & \texttt{\textbf{23.53}} & \texttt{\textbf{62.43}} \\
\texttt{GPT-5.1} & \texttt{39.53} & \texttt{8.99} & \texttt{19.67} & \texttt{62.33} \\
\bottomrule
\end{tabular}}
\caption{Best models on type-aware joint retrieval (RQ3), evaluated on comments with $\geq$2 evidence units. Full results in \S\ref{app:joint_retrieval}.}
\label{tab:joint_gold_label_retrieval_1}
\end{table}

Table~\ref{tab:joint_gold_label_retrieval_1} reports the best models on the 25.77\% of comments grounded to multiple evidence units, evaluated jointly under oracle type information. Performance drops sharply compared to the single-evidence oracle setting in Table~\ref{tab:oracle_text_retrieval}: even the strongest model (\texttt{Qwen-3 30B Instruct}) retrieves only 12.69\% of the full evidence set at R@1 and 23.53\% at R@2, despite the favorable type-aware conditions. This indicates that when evidence is distributed across non-redundant units, models struggle to surface them together.\footnote{Full results table and details in \S\ref{app:joint_retrieval}}

\subsection{Visual Evidence Retrieval (RQ4)}

Table~\ref{tab:visual_retrieval} shows that across all visual models, augmenting images with captions consistently improves image retrieval performance, indicating that captions provide a strong signal for grounding reviewer comments. Nonetheless, image-only retrieval remains competitive, hinting that visual information alone often encodes cues relevant to reviewer concerns (e.g., trends, comparisons, etc.). The remaining gap suggests that visual and textual signals are complementary, supporting RQ4 and motivating multimodal approaches for grounding.

\begin{table}[!t]
\centering
\small
\setlength{\tabcolsep}{3pt}
\renewcommand{\arraystretch}{1.30}
\resizebox{\columnwidth}{!}{
\begin{tabular}{lcccccc}
\toprule
 & \multicolumn{3}{c}{\textbf{Image+Caption}} 
 & \multicolumn{3}{c}{\textbf{Image-only}} \\
\cmidrule(lr){2-4} \cmidrule(lr){5-7}
\textbf{Model} 
& \textbf{MRR} & \textbf{R@1} & \textbf{R@2}
& \textbf{MRR} & \textbf{R@1} & \textbf{R@2} \\
\midrule
\multicolumn{7}{c}{\textbf{Table Retrieval}} \\
\texttt{SigLIP2} 
& \texttt{49.60} & \texttt{30.04} & \texttt{48.94}
& \texttt{44.54} & \texttt{24.40} & \texttt{42.56} \\
\texttt{OpenCLIP L14} 
& \texttt{47.23} & \texttt{28.48} & \texttt{44.82}
& \texttt{43.60} & \texttt{24.37} & \texttt{40.20} \\
\texttt{Jina Embeddings v4} 
& \texttt{55.67} & \texttt{37.19} & \texttt{55.92}
& \texttt{52.80} & \texttt{34.46} & \texttt{51.70} \\
\texttt{ColQwen2.5 v0.2} 
& \underline{\texttt{57.95}} & \underline{\texttt{40.48}} & \texttt{58.33}
& \texttt{53.73} & \texttt{35.65} & \texttt{52.16} \\
\texttt{Qwen-3 VL 8B} 
& \texttt{57.61} & \texttt{39.82} & \underline{\texttt{58.45}}
& \underline{\texttt{55.31}} & \underline{\texttt{37.87}} & \underline{\texttt{54.00}} \\
\texttt{Qwen-3 VL 32B} 
& \textbf{\texttt{59.53}} & \textbf{\texttt{42.39}} & \textbf{\texttt{60.34}}
& \textbf{\texttt{56.79}} & \textbf{\texttt{39.22}} & \textbf{\texttt{56.97}} \\
\midrule
\multicolumn{7}{c}{\textbf{Figure Retrieval}} \\
\texttt{SigLIP2} 
& \texttt{47.84} & \texttt{27.93} & \texttt{46.31}
& \texttt{44.23} & \texttt{24.02} & \texttt{42.53} \\
\texttt{OpenCLIP L14} 
& \texttt{46.99} & \texttt{26.54} & \texttt{46.26}
& \texttt{44.92} & \texttt{24.80} & \texttt{43.96} \\
\texttt{Jina Embeddings v4} 
& \texttt{52.11} & \texttt{32.88} & \texttt{52.87}
& \texttt{46.65} & \texttt{26.80} & \texttt{45.57} \\
\texttt{ColQwen2.5 v0.2} 
& \texttt{52.53} & \texttt{32.82} & \texttt{53.79}
& \texttt{46.54} & \texttt{26.10} & \texttt{46.14} \\
\texttt{Qwen-3 VL 8B} 
& \underline{\texttt{53.08}} & \underline{\texttt{33.49}} & \underline{\texttt{54.95}}
& \underline{\texttt{49.01}} & \underline{\texttt{28.93}} & \underline{\texttt{48.65}} \\
\texttt{Qwen-3 VL 32B} 
& \textbf{\texttt{56.39}} & \textbf{\texttt{37.66}} & \textbf{\texttt{58.47}}
& \textbf{\texttt{50.85}} & \textbf{\texttt{30.67}} & \textbf{\texttt{51.13}} \\
\bottomrule
\end{tabular}}
\caption{Visual evidence retrieval for tables and figures.}
\label{tab:visual_retrieval}
\end{table}

\section{Error Analysis}
\label{sec:error_analysis}
\subsection{Oracle text retrieval failures}
To understand how and why models are failing, we analyze \emph{oracle text retrieval} failures: for each type, we take the best model (\texttt{GPT-5.1} for paragraphs; \texttt{Qwen-3 30B Instruct} for sections/captions) and analyze 50 failed instances.

\textbf{For paragraphs} a dominant failure mode is due to line references being imprecise and pointing to section headers rather than the paragraph containing the relevant evidence. \textbf{For sections}, rebuttals often cite coarse sections while evidence is localized in a subsection. Moreover, line/section references frequently act as navigational pointers to another evidence unit (e.g. a table), so the text alone does not contain sufficient information to address the reviewer comment.

\textbf{Caption failures} are mainly due to some captions being too generic (with key details in the figure/table) or being similar across multiple figures/tables, causing plausible-but-wrong matches.

Across all types, \textbf{under-specified reviewer comments} that require additional local context (e.g., ''How did you introduce the error in Line~412?'') remain difficult to ground, as successful retrieval requires first identifying the referenced context before locating the explanatory evidence.

\subsection{Text vs.\ image retrieval disagreement}
Caption-based text retrieval and image-plus-caption-based retrieval often succeed on different instances. Looking at Recall@2, for figures, image retrieval is slightly higher than retrieving caption text only (58.47\% vs.\ 55.16\%; McNemar $p=0.25$) with substantial disagreement between both (both fail 26.0\%; disagree 24.1\%). For tables, image retrieval is higher and statistically significant (60.34\% vs.\ 53.29\%; $p=0.0156$) with high disagreement (both fail 24.8\%; disagree 29.7\%). Taken together, these findings show that reviewer comment grounding exhibits substantial modality-specific failure modes. Even when one modality is clearly stronger, neither text nor image retrieval alone is sufficient. A practical consequence is that systems grounding reviewer comments should consider late fusion strategies that combine caption-based and image-based scores, rather than committing to a single modality at retrieval time.

\section{Conclusion}
\label{sec:conclusion}

%We introduced a large-scale dataset for reviewer comment grounding that links reviewer comments to fine-grained, multimodal evidence in the original anonymous submission, using explicit references in author rebuttals as a high-precision annotation source. Experiments show that grounding reviewer comments is a difficult retrieval problem: performance drops sharply when models must both infer evidence type and retrieve relevant content, identifying type inference as a major bottleneck. Even under oracle conditions, semantic gaps between reviewer comments and relevant evidence limit retrieval effectiveness. Visual information improves grounding for figures and tables, but remains insufficient on its own, highlighting the need for multimodal fusion. Overall, our dataset exposes a realistic and demanding benchmark for scientific document understanding, and provides a foundation for future work on type-aware routing, multimodal retrieval, and multi-evidence grounding—key capabilities for reliable downstream tasks.

We introduce \texttt{\textbf{ReGround}}, a large-scale dataset that grounds reviewer comments in fine-grained, multimodal evidence from the original anonymous submission. We leverage author rebuttals as an annotation source: when authors point reviewers to a specific section, table, or figure, they indirectly annotate the link between a comment and the evidence that addresses it. This allowed us to scale to 3,656 papers and 16,274 reviewer comment–evidence pairs units while keeping the precision of expert human judgments. One limitation is that the target of retrieval is \textit{author-cited evidence}, so there might be plausible evidence not covered by that. 

Our grounding task presents a hard problem: unified retrieval peaks at 21\% Recall@10, type inference emerges as the dominant bottleneck, multi-evidence comments remain difficult even under oracle conditions, and text and image signals succeed on different instances making them complementary. These findings frame reviewer comment grounding as a stress test for scientific document understanding, demanding type-aware routing, multi-evidence aggregation, and multimodal fusion, capabilities that any system grounding claims in scientific documents will need.

\section*{Ethical considerations}
\label{sec:ethics}
This work uses peer-review data from NLPeer, consisting of original anonymous submissions, reviews, and rebuttals that were voluntarily donated by authors and reviewers, and publicly released under \texttt{CC-BY-NC 4.0}. All data were collected and processed in accordance with those terms, and will be released under the same \texttt{CC-BY-NC 4.0} license. Our dataset contains no personally sensitive information beyond what is publicly shared by the authors and reviewers. All annotators for the manual evaluation are volunteers. Our dataset is intended strictly for research on grounding reviewer comments in paper content, not for evaluating, scoring, or profiling individual reviewers or authors. While peer-review text may include subjective or critical language, our work does not aim to automate reviewer decisions or replace human judgment, but to support research on scientific document understanding and evidence retrieval.

\section*{Limitations}
\label{sec:limitations}

Our dataset focuses on grounding reviewer comments that are implicitly related to paper content through author rebuttals, and deliberately excludes comments that explicitly reference paper content--an easier but complementary setting. While this design choice isolates a more challenging and realistic grounding scenario, it does not capture the full spectrum of reviewer behaviors.

Our evaluation does not consider task-specific training or end-to-end systems that jointly perform evidence type selection and retrieval. We leave this for future work, and focus on analyzing the challenges in this work. In addition, dataset construction relies on paper references in author rebuttals, which reflect author interpretations of reviewer comments and emphasize content authors chose to address, potentially underrepresenting unresolved or weakly grounded comments. Moreover, author-provided evidence links can be coarse (e.g., section-level when a more specific subsection is appropriate), introducing unavoidable noise in the supervision signal, so retrieval performance should be interpreted with this limitation in mind.

Although we include multimodal evidence, text and image retrieval are evaluated using separate candidate pools; fully unified multimodal retrieval remains a non-trivial direction for future work. Finally, our dataset is limited to NLP papers due to inaccessibility of peer-review data from other domains. Other domains like biomedicine, geology or chemistry could have different evidence types and granularities not covered by this dataset. Investigating cross-domain reviewing dynamics is an important future direction, and we support this by releasing our dataset construction code publicly.

\section*{Acknowledgments}
This work has been co-funded by the German Federal Ministry of Research, Technology and Space (BMFTR) under the promotional reference 01ZZ2314H (GeMTeX), and by the European Union (ERC, InterText, 101054961). Views and opinions expressed are however those of the author(s) only and do not necessarily reflect those of the European Union or the European Research Council. Neither the European Union nor the granting authority can be held responsible for them. We thank Qian Ruan, Nils Dycke, and Dennis Zyska for their feedback on an initial draft of this paper.

\bibliography{anthology-1,anthology-2,custom}

\appendix

\section{Sentence Segmentation}
\label{app:sentence_splitter}

We segment reviews and rebuttals into sentence-level spans using a custom rule-based splitter. Off-the-shelf sentence segmenters frequently fail on peer-review text due to scientific abbreviations (e.g., `Fig.', `Sec.'), explicit manuscript references, and list-style formatting. Our splitter is designed to be deterministic and to preserve exact character offsets into the original text.

To prevent erroneous splits, we identify \emph{protected spans} in which punctuation should be ignored. These include explicit manuscript references detected using the patterns described in Appendix~\ref{app:rebuttal_references}, as well as common scientific abbreviations (e.g., `e.g.', `i.e.', `et al.'). Sentence-ending punctuation inside protected spans is excluded from boundary detection.

We additionally introduce \emph{forced sentence boundaries} at structural markers commonly used in reviews and rebuttals, including blank-line paragraph breaks, list items, block quotes, and headings. For headings followed by content on the same line (e.g., `Strengths:'), we split after the colon.

Outside protected spans, we split on sentence-ending punctuation characters (., !, ?) when followed by whitespace or the end of the text. We avoid splitting on numbered list markers (e.g., `1.' at line start) and include trailing closing characters such as quotes or parentheses with the preceding sentence.

Final sentence boundaries are obtained by merging punctuation-based split points with all forced boundaries. Each sentence is returned as a span with start and end character offsets into the original document, with leading and trailing whitespace trimmed. The complete implementation is released as part of our codebase.

\section{Detecting Explicit Rebuttal References}
\label{app:rebuttal_references}

This appendix describes how we detect explicit references to manuscript content in author rebuttals. These references form the supervision signal used to ground reviewer comments to fine-grained evidence in the original anonymous submission. Our goal in this step is to maximize precision while covering the broad range of reference formats used in scientific writing.

\subsection{Reference Types}
\label{app:reference_types}

We detect explicit references to the following categories of manuscript content:
(i) line numbers and line ranges,
(ii) sections and subsections,
(iii) figures,
(iv) tables,
(v) appendices,
(vi) equations,
(vii) pages, and
(viii) footnotes.
Table~\ref{tab:reference_types} summarizes the reference types along with representative examples observed in rebuttals.

\begin{table}[h]
\centering
\small
\begin{tabular}{l l}
\toprule
\textbf{Reference Type} & \textbf{Example Mentions} \\
\midrule
Lines & lines 120--135, l.~45--47, L120--L130 \\
Sections & Section~3.2, Sec.~4, §5, Results section \\
Figures & Figure~2, Fig.~3(a), Figs.~4--6 \\
Tables & Table~1, Tab.~5 \\
Appendices & Appendix~A, Appendix~C.2 \\
Equations & Eq.~(3), Equation~7 \\
Pages & page~5, pp.~3--4 \\
Footnotes & footnote~2, fn.~7 \\
\bottomrule
\end{tabular}
\caption{Types of explicit manuscript references detected in rebuttal sentences.}
\label{tab:reference_types}
\end{table}

\subsection{Regex-Based Reference Detection}
\label{app:regex_detection}

We identify reference-bearing rebuttal sentences using a set of regular expressions. These patterns were developed by manually inspecting 250 randomly sampled rebuttals and iteratively expanding coverage to capture common scientific reference conventions.

Patterns are grouped by reference type and account for common abbreviations (e.g., Fig. vs.\ Figure), optional punctuation, ranges (e.g., 2--4), and series (e.g., 2, 3, and 5). Matching is case-insensitive. A single rebuttal sentence may contain multiple references, in which case all references are extracted independently.

For readability, we describe the pattern families here and release the full implementation with our codebase.

\paragraph{Line references.}
We detect references to individual lines or line ranges, including formats such as lines 120--135, l.~45--47, and L120--L130. These patterns allow optional range markers (e.g., --, to) and optional parentheses.

\paragraph{Section references.}
We detect both numeric section references (e.g., Section~3.2, §4) and references to named section headers. Named section references include commonly used scientific sections such as \emph{Introduction}, \emph{Related Work}, \emph{Methodology}, \emph{Experiments}, \emph{Results}, \emph{Analysis}, \emph{Discussion}, \emph{Limitations}, and \emph{Conclusion}. These are matched using a curated list of canonical section names derived from ACL-style papers and empirical inspection of submissions. Matching is case-insensitive and allows optional determiners (e.g., the methodology section).

\paragraph{Figures and tables.}
We detect references to figures and tables using standard prefixes (e.g., Fig., Figure, Tab., Table), supporting single indices, ranges, and series (e.g., Figs.~2--4, Tables~1 and~3).

\paragraph{Appendices, equations, pages, and footnotes.}
We additionally detect references to appendices (e.g., Appendix~A.2), equations (e.g., Eq.~(5)), page ranges (e.g., pp.~3--4), and footnotes (e.g., footnote~7) using dedicated patterns for each category.

\subsection{Ambiguities and Resolution Strategy}
\label{app:reference_resolution}

When a rebuttal sentence contains multiple explicit references, we extract all references and generate multiple candidate links. During the subsequent content extraction step, each reference is resolved against the original anonymous submission PDF.

If a reference cannot be resolved (e.g., due to inconsistent line numbering, missing figures, or malformed indices), the corresponding instance is discarded. We do not attempt to infer or repair unresolved references in order to preserve the precision of the supervision signal.

References that primarily point to future revisions (e.g., ``we will update Figure~3 in the camera-ready version'') or to external papers are filtered out in a later stage, as described in \S\ref{app:implicit_filtering}.

All regular expressions and reference resolution utilities are implemented in Python and released as part of our public codebase.

\section{LLM-Based Rebuttal--Review Alignment}
\label{app:llm_alignment}

We align rebuttal sentences that contain explicit manuscript references to the reviewer comment they address.

\paragraph{Goal.}
Given a segmented reviewer report, a segmented rebuttal, and a target rebuttal sentence, the goal is to select the reviewer span(s) that the rebuttal sentence responds to. A rebuttal sentence may align to multiple reviewer spans.

\paragraph{Method.}
We perform alignment using \texttt{gpt-oss-120b} due to its long context window, and because it is one of the SOTA open-source models. The model is provided with (i) the full reviewer report with indexed spans, (ii) the full rebuttal text, and (iii) the target rebuttal sentence. It is instructed to return the identifiers of the reviewer span(s) being addressed, or an empty set if no clear alignment exists.

The full prompt template and output specification are shown in Figure~\ref{fig:alignment_prompt}. The model outputs span identifiers only for easier parsing.

\paragraph{Multiple and ambiguous cases.}
If a rebuttal sentence addresses multiple reviewer comments, all alignments are retained. Sentences that are generic or cannot be confidently aligned are discarded.

\section{Filtering}
\label{app:implicit_filtering}

To construct a dataset for \emph{implicit} reviewer comment grounding, we apply a set of precision-oriented filters to remove cases where the target evidence is explicitly identifiable or does not correspond to the reviewed submission. All filtering steps are applied after rebuttal--review alignment.

\subsection{Removing Quoted Review Text}

Authors frequently quote reviewer comments verbatim in rebuttals, which can introduce spurious references originating from the review rather than the author response. We remove rebuttal sentences that closely match reviewer text using fuzzy string matching.

\subsection{Removing Trivial Explicit-Reference Cases}
We remove instances where the reviewer comment already explicitly names the same evidence as the rebuttal (e.g., both reference the same figure, table, or section). This prevents trivial grounding via reference matching and ensures that the reviewer comment does not identify the target evidence.

\subsection{Removing References Outside the Reviewed Submission}
We remove rebuttal sentences that do not correspond to content available in the original anonymous submission. Specifically, we exclude cases where the rebuttal points to (i) promised camera-ready changes, (ii) new experiments or results introduced in the rebuttal, or (iii) content from external papers. This final step ensures that all retained pairs ground reviewer comments to content available in the original submission. This filter is implemented using a binary LLM; prompt template is in Fig.~\ref{fig:filtering_prompt}.

\section{Manual Validation Protocol}
\label{app:manual_validation}

We manually validated two stages of the dataset construction: (i) review--rebuttal alignment and (ii) filtering of invalid evidence.

\paragraph{Review--rebuttal alignment.}
Annotators were presented with the reviewer comment, the corresponding rebuttal text, and the automatically proposed alignment, with aligned spans highlighted. For each instance, annotators made a binary \emph{accept} or \emph{reject} decision. In cases of rejection, annotators provided a short keyword-based explanation (e.g., wrong reviewer comment, partial coverage, etc.). Annotators were not asked to provide corrected alignments.

\paragraph{Filtering validation.}
Filtering was validated using spreadsheet-based annotation. Each row corresponded to a single rebuttal sentence, and each column represented one filtering criterion (references to new experiments, future work, or external content). Annotators marked each criterion with a binary \emph{yes/no} decision. A rebuttal sentence was considered invalid evidence if it violated any filtering criterion.

\section{Manual Inspection}
\label{app:appendix_manual_validation}

Our manual inspection had two main goals.

First, we checked the most objective---and most failure-prone---component of the pipeline: reference resolution, i.e., whether the author-cited evidence correctly maps to the intended location in the paper. To assess this, one author manually inspected 200 reviewer comments containing a total of 316 references to paper evidence (restricted to text-only evidence). Our analysis shows that the vast majority of references are correctly resolved. Approximately 1.58\% (5/316) of the references may be considered coarse-grained, meaning that the cited span could potentially be narrowed to a more precise evidence boundary. However, determining appropriate evidence boundaries is inherently subjective, and in most cases it is difficult to definitively conclude that the boundaries selected by the authors are too broad.

Second, we checked the validity of the referenced evidence. We note that manually verifying whether a cited piece of evidence \textit{logically} addresses a reviewer comment is inherently subjective and requires deep understanding of paper-specific context. In practice, the authors of the paper are the most qualified annotators for this task. By relying on their rebuttals as a supervision signal, we effectively leverage the authors themselves as expert annotators of whether a given piece of evidence addresses a reviewer concern.

Nonetheless, for a subset of 100 reviewer comments, we checked each referenced evidence manually to determine their relevancy to the reviewer comment. We find that all referenced evidence is relevant to the reviewer comment. More importantly, in the majority of cases (84\%) the reviewer themselves acknowledged the rebuttal, which we take as a signal that the provided evidence was at least acceptable by the reviewer. Finally, our setup does not guarantee coverage: there might be other relevant evidence not referenced by the authors. However, we argue that this is unlikely, as authors would usually try to cover as much evidence as needed to strengthen their rebuttal

\begin{figure}[!t]
    \centering
    \includegraphics[width=\linewidth]{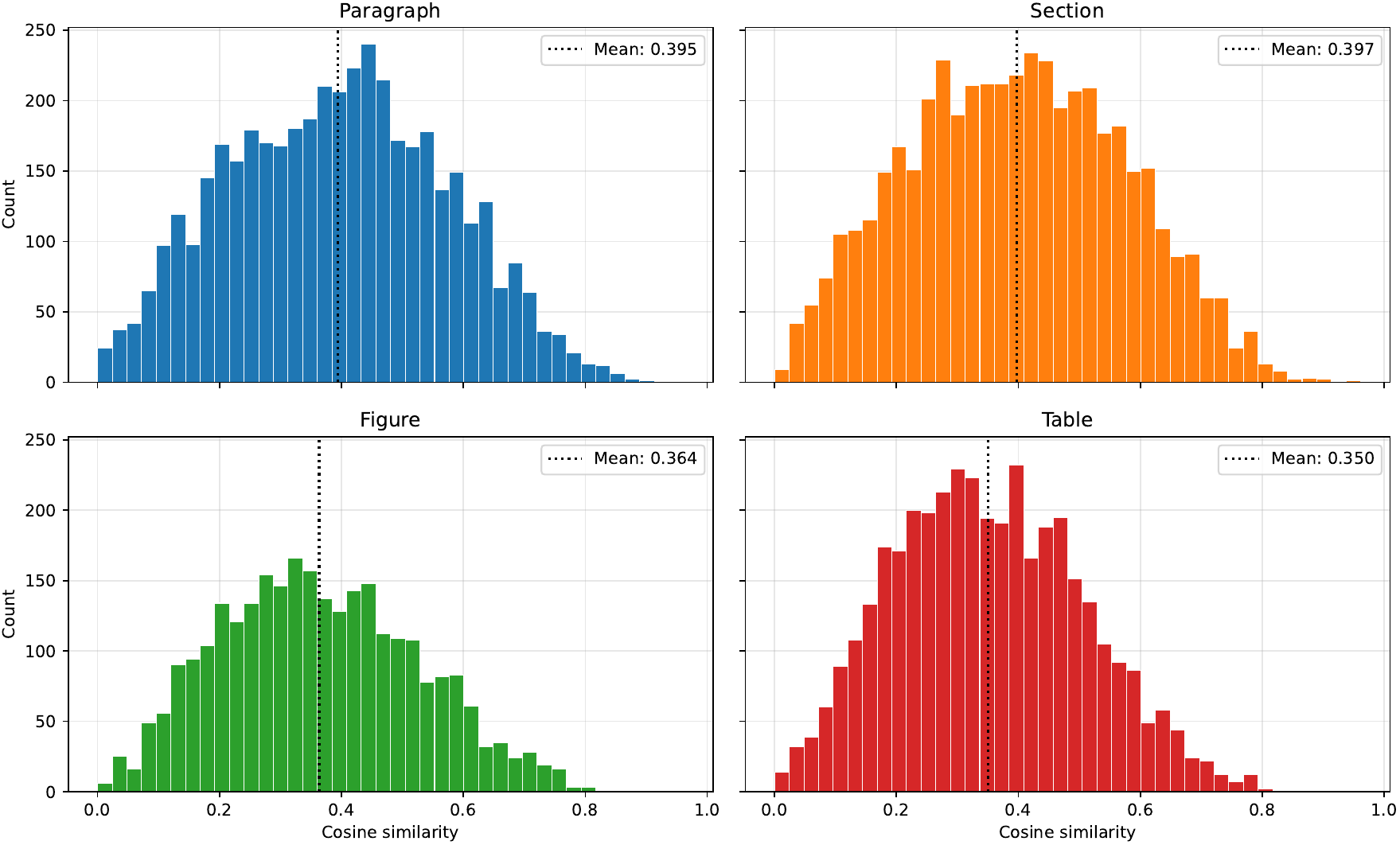}
    \caption{Distribution of cosine similarity between a reviewer comment and its target content unit(s).}
    \label{fig:similarity_dist}
\end{figure}

\section{Cosine Similarity Analysis}
\label{app:similarity}

We compute a simple embedding-based similarity measure between each reviewer comment and its grounded evidence unit(s) to characterize their lexical/semantic proximity.

\paragraph{Embedding model and similarity.}
For each comment--evidence pair, we encode the reviewer comment and the evidence text using the Sentence-Transformers model \texttt{all-mpnet-base-v2}. We then compute cosine similarity between the two embeddings. For figure/table evidence, we use the associated caption text as the evidence text. Similarities are computed independently per grounded pair, and pairs with multiple evidence units contribute one similarity value per unit.

\paragraph{Distributions by evidence type.}
Figure~\ref{fig:similarity_dist} shows histograms of cosine similarity values stratified by reference type (e.g., lines, sections, figures, tables). Across the dataset, the mean cosine similarity over all reviewer comment--evidence pairs is 0.377. We observe substantial overlap across types, with no evidence category exhibiting consistently high similarity scores. 

\noindent
We report these statistics solely as a descriptive property of the dataset and do not use them as supervision or a filtering criterion.

\section{Multi-Evidence Type Combinations}
\label{app:evidence-combinations}

Comments grounded to multiple evidence units are particularly challenging: retrieval must identify and aggregate all relevant units rather than retrieve a single passage. To characterize this setting, Table~\ref{tab:evidence-combinations} breaks down the most frequent and substantively informative evidence-type combinations at the query level.  Percentages are calculated within the multi-evidence subset. The displayed combinations account for 55\% of all multi-evidence combinations.

\begin{table}[t]
\centering
\small
\begin{tabular}{lr}
\toprule
\textbf{Combination} & \textbf{\% of multi-evidence comments} \\
\midrule
Paragraph + section & 14.06 \\
Multiple paragraphs & 8.84 \\
Multiple sections & 7.79 \\
Section + table & 6.31 \\
Figure + section & 5.86 \\
Figure + table & 5.14 \\
Multiple tables & 4.54 \\
Paragraph + table & 3.17 \\
\bottomrule
\end{tabular}
\caption{Most frequent evidence-type combinations for comments grounded to
multiple evidence units. Percentages are within the multi-evidence subset.}
\label{tab:evidence-combinations}
\end{table}

The challenge is therefore not confined to a single aggregation setting. The dataset includes same-type aggregation (e.g., multiple paragraphs, sections, or tables), cross-granularity aggregation (e.g., paragraph + section), and multimodal combinations involving figures or tables. This diversity makes multi-evidence retrieval a test of both finding individual units and combining evidence across content types.

\section{Relation Directionality}
\label{app:relation-directionality}

ReGround formalizes retrieval of the paper content to which a reviewer comment
and an author response refer.  The direction of the relation is a separate
question from retrieval: relevant evidence can be retrieved before determining
whether it supports, qualifies, or contradicts the reviewer comment.  Thus,
our use of \emph{evidence} does not imply a verdict or relation type.

As a descriptive check for data analysis, one annotator manually labeled a sample of 200
reviewer-comment--rebuttal-sentence pairs using a narrower three-way scheme.
A relation was labeled \emph{corrective} when the authors used cited paper
content to contradict a reviewer claim or show that allegedly missing content
was already present.  It was labeled \emph{explanatory} when the cited content
answered, clarified, justified, or otherwise addressed a concern without
establishing that the reviewer was mistaken; the remaining cases were labeled
\emph{other/unclear}.  In this sample, 126 pairs (63.0\%) were explanatory,
64 (32.0\%) corrective, and 10 (5.0\%) other/unclear.

These single-annotator descriptive results should be interpreted cautiously,
but they suggest that corrective uses of evidence are not the dominant pattern
in the sample.  More commonly, authors point to paper content to explain or
clarify a concern.  We therefore retain the retrieval framing and leave
relation typing as a distinct downstream classification layer: evidence must
be retrieved before its relation type to the reviewer comment can be assessed.

\begin{table}[!t]
\centering
\setlength{\tabcolsep}{6pt}
\begin{tabular}{l c}
\toprule
\textbf{Primary Dimension} & \textbf{Percentage} \\
\midrule
Empirical Rigor & 30.8\% \\
Technical Soundness & 15.7\% \\
Deployment \& Impact & 15.6\% \\
Presentation \& Writing & 12.8\% \\
Data \& Reproducibility & 11.5\% \\
Conceptual \& Novelty & 11.1\% \\
Other & 2.5\% \\
\bottomrule
\end{tabular}
\caption{Distribution of primary reviewer comment dimensions across the annotated examples, based on taxonomy classification.}
\label{tab:taxonomy_dist}
\end{table}

\section{Reviewer Comment Taxonomy}
\label{app:taxonomy}

\paragraph{Annotation setup.}
Each comment is assigned a single primary category corresponding to the dominant aspect of the paper discussed. Categories are defined to be mutually exclusive and collectively exhaustive. The resulting taxonomy is intended to be descriptive rather than normative. The annotation is done by one of the authors. Table~\ref{tab:taxonomy_dist} reports the resulting distribution over primary categories.

\paragraph{Taxonomy categories.}
The taxonomy consists of the following categories:
(i) \textbf{Conceptual \& Novelty}: originality, significance, and relation to prior work;
(ii) \textbf{Technical Soundness}: correctness of theoretical claims and technical reasoning;
(iii) \textbf{Empirical Rigor}: quality of experiments, evaluations, and analyses;
(iv) \textbf{Data \& Reproducibility}: data quality, annotation procedures, and reproducibility concerns;
(v) \textbf{Presentation \& Writing}: clarity, organization, and visual presentation;
(vi) \textbf{Deployment \& Impact}: efficiency, ethics, and real-world applicability;
(vii) \textbf{Other}: administrative, meta-level, or vague comments not fitting elsewhere.

\begin{table*}[!t]
\centering
\small
\begin{tabular}{lrrr}
\toprule
\textbf{Dimension} & \textbf{Included} & \textbf{Excluded} & \textbf{Difference} \\
\midrule
Empirical Rigor & 30.8\% & 29.5\% & +1.3 \\
Technical Soundness & 15.7\% & 9.0\% & +6.7 \\
Deployment \& Impact & 15.6\% & 3.0\% & +12.6 \\
Presentation \& Writing & 12.8\% & 14.0\% & $-1.2$ \\
Data \& Reproducibility & 11.5\% & 10.0\% & +1.5 \\
Conceptual \& Novelty & 11.1\% & 16.5\% & $-5.4$ \\
Other & 2.5\% & 18.0\% & $-15.5$ \\
\bottomrule
\end{tabular}
\caption{Comment-type distribution in ReGround and 200 excluded comments.}
\label{tab:coverage-types}
\end{table*}

\section{Coverage and Representativeness Analysis}
\label{app:coverage}

Our rebuttal-based construction necessarily selects reviewer comments that authors address by referring to content in the paper.  To quantify this selection, we measure coverage over complete review threads using the same sentence-level segmentation as the construction pipeline.  Because rebuttals primarily address criticisms and suggestions, rather than paper summaries or strengths, we compute coverage only over the \texttt{summary\_of\_weaknesses} and \texttt{comments\_suggestions\_and\_typos} fields.  In this subset, ReGround covers 17,682 of 115,989 review spans (15.24\%).

We also assess whether the included comments differ systematically from those excluded by this procedure.  We randomly sampled 200 excluded spans and classified them using the same seven-category taxonomy and definitions used elsewhere in our analysis.  Table~\ref{tab:coverage-types} compares their distribution with that of included comments. 

The largest difference is in \emph{Other} comments, which comprise administrative, meta-level, or vague remarks.  Such comments generally do not identify paper content to retrieve; their exclusion is therefore a property of the grounding task rather than evidence of a content-related sampling bias. The same partly applies to \emph{Conceptual \& Novelty} comments, which often concern positioning relative to the field rather than a localized part of the paper. Conversely, ReGround over-represents concerns that authors can answer by pointing to paper content, particularly \emph{Technical Soundness} and \emph{Deployment \& Impact}. These are precisely the comments for which a grounding system is intended to be used.  Importantly, \emph{Empirical Rigor}, the largest category, has nearly identical prevalence among included and excluded comments (30.8\% versus 29.5\%), indicating no material skew in the dominant reviewer concern.

\begin{table*}[!t]
\centering
\resizebox{\textwidth}{!}{
\begin{tabular}{lcccccccccc}
\hline
\textbf{Model} & \textbf{Q1} & \textbf{Q2} & \textbf{Q3} & \textbf{Q4} & \textbf{Q5} & \textbf{Q6} & \textbf{Q7} & \textbf{Q8} & \textbf{Q9} & \textbf{Q10} \\
\hline
\texttt{BGE-M3-Reranker} & \texttt{21.0} & \texttt{29.9} & \texttt{34.2} & \texttt{41.5} & \texttt{44.2} & \texttt{39.2} & \texttt{38.8} & \texttt{31.6} & \texttt{12.7} & \texttt{0.2} \\
\texttt{Qwen3-30B-Instruct} & \texttt{18.2} & \texttt{32.3} & \texttt{33.8} & \texttt{39.0} & \texttt{41.8} & \texttt{45.0} & \texttt{44.9} & \texttt{46.1} & \texttt{30.4} & \texttt{7.2} \\
\hline
\end{tabular}}
\caption{MRR across section length quantiles for the best cross-encoder (BGE-M3-Reranker) and the best LLM-based retriever (Qwen3-30B-Instruct).}
\label{tab:length_quantiles}
\end{table*}

\section{LLM-based Pointwise Ranking}
\label{app:llm_ranking}

\paragraph{Setup.}
We use LLMs as pointwise relevance judges. Given a reviewer comment $c$ and a candidate evidence unit $d$ (paragraph/section/caption), the model predicts whether $d$ helps address $c$. Each candidate is scored independently; candidates are then ranked by this score.

\paragraph{Scoring from token log-probabilities.}
We request token-level log-probabilities (\texttt{logprobs}) and a small set of \texttt{top\_logprobs} for each generated token. Let $\ell_A$ and $\ell_B$ denote the log-probabilities of emitting \texttt{A} and \texttt{B} at the final decision position. We compute a calibrated relevance score as the normalized probability of \texttt{A}:
\begin{equation}
p(\texttt{A}\mid c,d)
= \frac{\exp(\ell_A)}{\exp(\ell_A)+\exp(\ell_B)}.
\end{equation}
This score is used for ranking (higher is more relevant). If only one of $\ell_A,\ell_B$ is present in \texttt{top\_logprobs}, we fall back to the emitted token's own log-probability when it matches \texttt{A} or \texttt{B}.

\paragraph{Decision token identification.}
Because tokenizers may emit variants such as \texttt{" A"} vs.\ \texttt{"A"}, we normalize tokens by stripping whitespace. We select the decision index by (i) taking the last non-whitespace character in the generated text and matching it to \texttt{A}/\texttt{B}, then (ii) falling back to the last generated token whose normalized form is in \{\texttt{A},\texttt{B}\}. If no decision token is found (formatting violation), we retry the call up to a fixed number of attempts and otherwise mark the instance as failed.

\paragraph{Inference settings and model-specific controls.}
All calls use the same system message and user prompt template. We set decoding parameters per model family to reduce formatting failures and support explicit ``thinking'' modes when available. Concretely, we use temperature $0$ for non-thinking settings with a small \texttt{max\_tokens} budget (enough to emit only the label), and enable model-provided thinking modes where supported (e.g., via template flags) with a higher temperature. We request \texttt{top\_logprobs} to increase the chance that both labels appear in the returned distribution.

\paragraph{Failure handling.}
We treat three cases as failures: missing token log-probabilities, inability to locate the decision token, or neither label appearing in the returned \texttt{top\_logprobs}. We retry formatting-related failures up to a fixed maximum.

\section{Effect of Evidence Length}
\label{app:length_effect}
To analyze the effect of section length on retrieval performance, we evaluate the Mean Reciprocal Rank (MRR) across ten quantiles of section text length. Quantile Q1 corresponds to the shortest sections and Q10 to the longest.

The results in Table~\ref{tab:length_quantiles} show different behaviors for the two model types. The cross-encoder (BGEM3) improves as section length increases up to mid-length sections (Q4–Q5), after which performance declines sharply, with a substantial drop for the longest sections. This supports our hypothesis that cross-encoders struggle when ranking very long text segments.

In contrast, the LLM-based retriever (Qwen3-30B-Instruct) continues to improve with increasing section length up to Q8, only decreasing in the final two quantiles. This behavior is expected, as longer sections provide more contextual information that the model can use to determine whether a section is relevant to a reviewer comment.

\begin{table}[!t]
\centering
\small
\renewcommand{\arraystretch}{1}
\resizebox{\linewidth}{!}{
\begin{tabular}{lcccc}
\toprule
\textbf{Model} & \textbf{MRR} & \textbf{R@1} & \textbf{R@2} & \textbf{R@10} \\
\midrule
\textit{\textbf{Sparse encoders}} & & & & \\
\texttt{BM25} & \texttt{29.91} & \texttt{6.88} & \texttt{13.31} & \texttt{46.78} \\
\texttt{SPLADEv3} & \texttt{37.17} & \texttt{10.28} & \texttt{18.70} & \texttt{54.51} \\
\midrule
\textit{\textbf{Dense encoders}} & & & & \\
\texttt{all-mpnet-base-v2} & \texttt{32.54} & \texttt{7.82} & \texttt{15.08} & \texttt{49.47} \\
\texttt{BGE-M3} & \texttt{34.85} & \texttt{9.18} & \texttt{16.62} & \texttt{51.75} \\
\texttt{Qwen-3 Embedding 4B} & \texttt{26.29} & \texttt{7.46} & \texttt{13.06} & \texttt{38.11} \\
\texttt{EmbeddingGemma} & \texttt{37.63} & \texttt{10.25} & \texttt{18.79} & \texttt{55.94} \\
\midrule
\textit{\textbf{Cross encoders}} & & & & \\
\texttt{BGE-M3-Reranker} & \texttt{36.91} & \texttt{9.93} & \texttt{17.99} & \texttt{54.80} \\
\texttt{MiniLM-L12-v2} & \texttt{33.88} & \texttt{8.98} & \texttt{16.08} & \texttt{50.30} \\
\midrule
\textit{\textbf{LLMs}} & & & & \\
\texttt{Gemma-3 12B} & \texttt{39.82} & \texttt{10.54} & \texttt{21.46} & \texttt{59.32} \\
\texttt{Gemma-3 27B} & \texttt{43.20} & \texttt{12.59} & \textbf{\texttt{23.91}} & \texttt{62.21} \\
\texttt{Qwen-3 30B Instruct} & \textbf{\texttt{43.29}} & \textbf{\texttt{12.69}} & \texttt{23.53} & \textbf{\texttt{62.43}} \\
\texttt{Qwen-3 30B Thinking} & \texttt{38.34} & \texttt{11.14} & \texttt{18.44} & \texttt{54.53} \\
\texttt{gpt-oss-20b} & \texttt{39.44} & \texttt{11.69} & \texttt{19.01} & \texttt{52.95} \\
\texttt{GPT-5.1} & \texttt{39.53} & \texttt{8.99} & \texttt{19.67} & \texttt{62.33} \\
\bottomrule
\end{tabular}}
\caption{Joint retrieval over all gold labels for a query, restricted to the gold evidence types.}
\label{tab:joint_gold_label_retrieval}
\end{table}

\begin{table*}[!t]
\centering
\small
\setlength{\tabcolsep}{4pt}
\resizebox{\textwidth}{!}{
\begin{tabular}{llrrrrrrrrrr}
\toprule
\textbf{Setting} & \textbf{Predictor} & \textbf{Set Acc.} & \textbf{Micro-F1} & \textbf{Macro-F1} & \textbf{Avg. Gold} & \textbf{Avg. Pred.} & \textbf{Fig. F1} & \textbf{Tab. F1} & \textbf{Text F1} & \textbf{Par. F1} & \textbf{Sec. F1} \\
\midrule
3-way & \texttt{EmbeddingGemma} & 30.28 & 60.20 & 49.36 & 1.07 & 1.62 & 33.82 & 37.15 & 77.12 & -- & -- \\
3-way & \texttt{Qwen-3 30B Instruct} & 55.98 & 61.43 & 34.31 & 1.07 & 1.04 & 12.82 & 12.72 & 77.41 & -- & -- \\
3-way & \texttt{GPT-5.1} & 50.12 & 61.51 & 42.33 & 1.07 & 1.19 & 20.58 & 29.48 & 76.95 & -- & -- \\
\midrule
4-way & \texttt{EmbeddingGemma} & 3.25 & 46.36 & 45.33 & 1.14 & 2.46 & 37.50 & 40.53 & -- & 51.62 & 51.69 \\
4-way & \texttt{Qwen-3 30B Instruct} & 29.15 & 39.39 & 27.29 & 1.13 & 1.11 & 17.99 & 16.27 & -- & 18.84 & 56.07 \\
4-way & \texttt{GPT-5.1} & 19.14 & 42.88 & 38.58 & 1.14 & 1.54 & 24.26 & 38.37 & -- & 39.04 & 52.64 \\
\bottomrule
\end{tabular}}
\caption{Standalone multi-label evidence-type prediction performance. Scores are percentages except for the average number of gold/predicted labels. In the 3-way setting, \textit{text} merges paragraph- and section-level evidence.}
\label{tab:type_prediction}
\end{table*}

\begin{table*}[!t] \centering \small \setlength{\tabcolsep}{6pt} \resizebox{\textwidth}{!}{ \begin{tabular}{llrrrr} \toprule \textbf{Setting} & \textbf{Retriever} & \textbf{Base R@10} & \textbf{Filtered R@10} & \textbf{Base MRR} & \textbf{Filtered MRR} \\ \midrule 3-way & \texttt{GPT-5.1} & 21.67 & $15.92_{\scriptscriptstyle -5.75}$ & 10.92 & $7.84_{\scriptscriptstyle -3.08}$ \\ 3-way & \texttt{Qwen-3 30B Instruct} & 21.15 & $14.57_{\scriptscriptstyle -6.58}$ & 10.87 & $6.68_{\scriptscriptstyle -4.19}$ \\ 3-way & \texttt{EmbeddingGemma} & 17.68 & $16.41_{\scriptscriptstyle -1.27}$ & 8.78 & $9.09_{\scriptscriptstyle +0.31}$ \\ \midrule 4-way & \texttt{GPT-5.1} & 21.67 & $15.02_{\scriptscriptstyle -6.65}$ & 10.92 & $7.18_{\scriptscriptstyle -3.74}$ \\ 4-way & \texttt{Qwen-3 30B Instruct} & 21.15 & $6.11_{\scriptscriptstyle -15.04}$ & 10.87 & $3.05_{\scriptscriptstyle -7.82}$ \\ 4-way & \texttt{EmbeddingGemma} & 17.68 & $18.16_{\scriptscriptstyle +0.48}$ & 8.78 & $9.71_{\scriptscriptstyle +0.93}$ \\ \bottomrule \end{tabular}} \caption{Retrieval after hard filtering by predicted evidence type on the shared subset. Deltas in subscripts indicate the change from the base retriever. Filtering usually lowers R@10 and MRR because type prediction errors remove relevant evidence before ranking.} \label{tab:predicted_type_retrieval} \end{table*}

\section{Joint Retrieval Remains Challenging for Multi-Evidence Comments}
\label{app:joint_retrieval}
Table~\ref{tab:joint_gold_label_retrieval} reports joint retrieval results for reviewer comments grounded in multiple evidence units, assuming oracle evidence types. Even in this favorable setting, performance remains limited, showing that multi-evidence grounding poses challenges beyond type inference.

Across all models, recall at small cutoffs (R@1 and R@2) is substantially lower than in the single-target oracle setting, showing that models struggle to surface multiple relevant evidence units among the top-ranked results equally. While larger LLMs achieve higher overall recall, the gains are modest, and no model reliably retrieves the full set of relevant evidence early in the ranking.

These results support RQ3 and suggest that reviewer comments often require aggregating complementary evidence that is distributed across the paper. Even when the correct evidence types are known, retrieving all relevant units within a single ranked list remains difficult, highlighting multi-evidence retrieval as a distinct challenge in reviewer comment grounding.

\section{Predicted Evidence-Type Routing}
\label{app:predicted_type_routing}

The oracle type-aware setting in Section~\ref{sec:results} assumes access to the gold evidence type. To test whether this information can be inferred automatically, we run a two-stage experiment: first predict the relevant evidence type(s) from the reviewer comment, then retrieve only from candidates of the predicted type(s). We evaluate a four-way setting (figure, table, paragraph, section) and a coarser three-way setting (figure, table, text), where \textit{text} includes both paragraph- and section-level evidence. Table~\ref{tab:type_prediction} reports the type-prediction quality, and Table~\ref{tab:predicted_type_retrieval} reports retrieval performance on the shared subset.

Table~\ref{tab:type_prediction} shows that predictors identify the dominant \textit{text} class reliably, but figure and table prediction remains substantially weaker. This explains the retrieval behavior in Table~\ref{tab:predicted_type_retrieval}: the broad \textit{text} class leaves many candidates in the pool, while mistakes on figure/table examples remove the correct evidence entirely. Thus, predicted evidence type is better interpreted as a soft reranking signal than as a hard retrieval constraint.

\section{AI Assistants Usage}
We used GitHub Copilot for some coding related tasks, as well as ChatGPT for light editing (phrasing, grammar proof-checking) to help writing the paper.

\begin{figure*}[!h]   
\centering
\begin{tcolorbox}[colback=gray!5, colframe=gray!80, fontupper=\ttfamily, title=Review-Rebuttal Alignment Prompt, width=\linewidth]
\footnotesize
Your task is to align a rebuttal sentence to the corresponding reviewer spans it's replying to.

Read the rebuttal carefully first.\\
The relevant reviewer point is usually indicated earlier in the rebuttal text, often by:\\
- direct or partial quoting of the relevant reviewer span,\\
- explicit markers like "Weakness 2", "W2", "Q3", "Comment 1", "R1", etc.,\\
- paraphrased references to the reviewer's wording.\\
Use these signals if present.

\#\#\# Reviewer Segments (Candidates)\\
Choose ONLY from these IDs. Do NOT create text, do NOT rewrite spans.
\{spans\_str\}

\#\#\# Full Rebuttal\\
\{rebuttal\_text\}

\#\#\# Target Rebuttal Sentence\\
"\{target\_sentence\}"

\#\#\# Instructions\\
1. Determine which reviewer segment(s) the target rebuttal sentence responds to.\\
2. If multiple points are being addressed, return all relevant segments.

\#\#\# Output\\
Return ONLY JSON in the following format:\\
\{\{"review\_segments": ["sent\#1", "sent\#2"]\}\}

Do not include explanations, comments, or markdown fences.\\
\end{tcolorbox}
\caption{Prompt template for review--rebuttal alignment.}
\label{fig:alignment_prompt}
\end{figure*}

\begin{figure*}[t]   
\centering
\begin{tcolorbox}[colback=gray!5, colframe=gray!80, fontupper=\ttfamily, title=LLM Filtering Prompt, width=\linewidth]
\footnotesize
You are analyzing one rebuttal sentence from an academic peer-review exchange.\\
Your goal is to decide whether this sentence explicitly refers to content from the ORIGINAL ANONYMOUS SUBMISSION, or whether it should be excluded for one of several reasons. Read ALL provided text, but focus your decision ONLY on the TARGET REBUTTAL SENTENCE.

------------------------------------------------------------\\
REVIEW TEXT:\\
\{review\_text\}

------------------------------------------------------------\\
REBUTTAL TEXT:\\
\{rebuttal\_text\}

------------------------------------------------------------\\
RELEVANT REVIEW COMMENT:\\
\{rebuttal\_comment\}

------------------------------------------------------------\\
TARGET REBUTTAL SENTENCE (the ONLY sentence to classify):\\
"\{target\_sentence\}"\\
------------------------------------------------------------

Answer the following THREE questions. Each must be strictly true or false.\\
Interpret each question EXACTLY as defined.

\#\#\# Q1: Future Changes / Camera-Ready\\
Does the sentence describe something the authors *plan* to add, modify, remove, clarify, expand, or correct in the future, including in the camera-ready version or in any revised version?

\#\#\# Q2: New Experiments or Newly Introduced Results (During Rebuttal)\\
Does the sentence report or describe new experiments, analyses, tables, figures, metrics, or results that were NOT part of the original anonymous submission?

\#\#\# Q3: External Papers / Citations\\
Does the sentence explicitly reference external papers, authors, or citations in a way that is not part of describing content in the original submission?

------------------------------------------------------------\\
\#\#\# REQUIRED OUTPUT FORMAT\\
Return ONLY valid JSON with ALL fields present:

\{\{
  "future\_changes": true/false,\\
  "new\_experiments": true/false,\\
  "external\_references": true/false\\
\}\}

Do NOT provide explanations. Do NOT include text outside the JSON. Return ONLY the JSON object.
\end{tcolorbox}
\caption{Prompt template for LLM-based pairwise sentence classification.}
\label{fig:filtering_prompt}
\end{figure*}

\end{document}